\documentclass[letterpaper, 10 pt, conference]{IEEEconf}  
\pdfoutput=1

\IEEEoverridecommandlockouts   
\usepackage{graphicx} 

\usepackage[T1]{fontenc}
\usepackage{epsfig} 
\usepackage{etex}
\usepackage{amsmath} 
\usepackage{amssymb}  
\usepackage{color} 
\usepackage{mathtools}
\usepackage{booktabs}
\usepackage{times}
\usepackage[labelfont=bf, font=footnotesize]{caption}
\usepackage[labelfont=bf, font=footnotesize]{subcaption}

\usepackage{enumerate}
\usepackage{balance}
\usepackage{algorithm}
\usepackage{setspace}
\usepackage{varwidth}
\usepackage{algpseudocode}
\usepackage{textcomp}
\usepackage{gensymb}
\usepackage{tabulary}
\usepackage[numbers,sort]{natbib}
\usepackage{dblfloatfix}
\usepackage{relsize}
\usepackage{graphbox} 
\usepackage{import}
\usepackage{tabularx}
\usepackage{array}
\usepackage{makecell}
\usepackage{multirow}
\usepackage{dblfloatfix}  
\usepackage{float}
\usepackage{xargs} 
\usepackage[table, pdftex,dvipsnames]{xcolor}  
\usepackage[export]{adjustbox}
\usepackage{hyperref}
\usepackage{url}            
\let\labelindent\relax 
\usepackage{enumitem}
\usepackage{blindtext}
\usepackage{regexpatch}
\usepackage{tikz} 
\graphicspath{{images/}}

\usepackage{soul}
\colorlet{suddhu}{BurntOrange!30}
\colorlet{michael}{CornflowerBlue!30}
\colorlet{alberto}{ForestGreen!30}
\colorlet{maria}{WildStrawberry!30}
\colorlet{peter}{Emerald!30}
\colorlet{josh}{Fuchsia!30}
\newboolean{draft}
\setboolean{draft}{true}
\DeclareRobustCommand{\suddhu}[1]{\ifthenelse{\boolean{draft}}{\begingroup\sethlcolor{suddhu}\hl{(Suddhu) #1}\endgroup  \\} }
\DeclareRobustCommand{\michael}[1]{\ifthenelse{\boolean{draft}}{\begingroup\sethlcolor{michael}\hl{(Michael) #1}\endgroup \\} }
\DeclareRobustCommand{\alberto}[1]{\ifthenelse{\boolean{draft}}{\begingroup\sethlcolor{alberto}\hl{(Alberto) #1}\endgroup \\} }
\DeclareRobustCommand{\maria}[1]{\ifthenelse{\boolean{draft}}{\begingroup\sethlcolor{maria}\hl{(Maria) #1}\endgroup \\} }
\DeclareRobustCommand{\peter}[1]{\ifthenelse{\boolean{draft}}{\begingroup\sethlcolor{peter}\hl{(Peter) #1}\endgroup \\} }
\DeclareRobustCommand{\josh}[1]{\ifthenelse{\boolean{draft}}{\begingroup\sethlcolor{josh}\hl{(Josh) #1}\endgroup \\} }

\hypersetup{
    colorlinks=true,
    filecolor=magenta,      
    urlcolor=magenta,
    citecolor=black,
    pdftitle=Tactile SLAM: Real-time inference of shape and pose from planar pushing,
    pdfauthor={Sudharshan Suresh, Maria Bauza, Kuan-Ting Yu, Joshua G. Mangelson, Alberto Rodriguez, Michael Kaess}
}
\definecolor{qcolor}{HTML}{FFA700}
\definecolor{icolor}{HTML}{075493}
\definecolor{pcolor}{HTML}{76D6FF}
\definecolor{fcolor}{HTML}{942193}
\definecolor{prcolor}{HTML}{929292}

\definecolor{contact_color}{HTML}{0B1BE2}
\definecolor{motion_color}{HTML}{FEA700}
\definecolor{shape_color}{HTML}{25A730}
\definecolor{shape_fill_color}{HTML}{D2E1D4}

\algtext*{EndWhile}
\algtext*{EndIf}
\algtext*{EndFor}

\newcommand*\mystrut[1]{\vrule width0pt height0pt depth#1\relax}
\newcommand{\etal}{et al.}

\newcommand{\GobbleSmall}[0]{\vspace{-0.5\baselineskip}}
\newcommand{\GobbleMedium}[0]{\vspace{-1.0\baselineskip}}
\newcommand{\GobbleLarge}[0]{\vspace{-1.5\baselineskip}}

\newcommand{\raisemath}[1]{\mathpalette{\raisem@th{#1}}}
\newcommand{\boldsubheading}[1]{\vspace{0.1in}\noindent\textbf{#1:}}

\title{\LARGE \bf Tactile SLAM: Real-time inference of shape and pose \\ from planar pushing}

\author{Sudharshan Suresh$^1$, Maria Bauza$^2$, Kuan-Ting Yu$^3$, \\ Joshua G. Mangelson$^4$, Alberto Rodriguez$^2$, and Michael Kaess$^1$
\thanks{$^{1}$Sudharshan Suresh and Michael Kaess are with the Robotics Institute, Carnegie Mellon University {\tt\footnotesize <suddhu,kaess>@cmu.edu}}%
\thanks{$^{2}$Maria Bauza and Alberto Rodriguez are with the Mechanical Engineering Department, Massachusetts Institute of Technology {\tt\footnotesize <bauza,albertor>@mit.edu}}%
\thanks{$^{3}$Kuan-Ting Yu is with XYZ Robotics {\tt\footnotesize peterkty@gmail.com}}%
\thanks{$^{4}$Joshua G. Mangelson is with the Electrical and Computer Engineering Department, Brigham Young University {\tt\footnotesize joshua\_mangelson@byu.edu}}%
\thanks{This work was partially supported by the National Science Foundation under award IIS-2008279. We thank Eric Dexheimer, Daniel McGann, Allison Wong, and Paloma Sodhi for their discussions and feedback.}
\thanks{\hrule}
\thanks{Project website and code: \href{http://www.cs.cmu.edu/\textasciitilde sudhars1/tactile-slam/}{cs.cmu.edu/\textasciitilde sudhars1/tactile-slam/}}
}



\begin{document}

\maketitle
\thispagestyle{empty}
\pagestyle{empty}

\begin{abstract}
Tactile perception is central to robot manipulation in unstructured environments. However, it requires contact, and a mature implementation must infer object models while also accounting for the motion induced by the interaction. In this work, we present a method to estimate both object shape and pose in real-time from a stream of tactile measurements. This is applied towards tactile exploration of an unknown object by planar pushing. We consider this as an online SLAM problem with a nonparametric shape representation. Our formulation of tactile inference alternates between Gaussian process implicit surface regression and pose estimation on a factor graph. Through a combination of local Gaussian processes and fixed-lag smoothing, we infer object shape and pose in real-time. We evaluate our system across different objects in both simulated and real-world planar pushing tasks. 
\end{abstract}
\GobbleSmall

\section{Introduction}
\label{sec:intro}

For effective interaction, robot manipulators must build and refine their understanding of the world through sensing. This is especially relevant in unstructured settings, where robots have little to no knowledge of object properties, but can physically interact with their surroundings. Even when blindfolded, humans can locate and infer properties of unknown objects through touch~\cite{klatzky1985identifying}. Replicating some fraction of these capabilities will enable contact-rich manipulation in environments such as homes and warehouses. In particular, knowledge of object shape and pose determines the success of generated grasps or nonprehensile actions. 


While there have been significant advances in tactile sensing, from single-point sensors to high-resolution tactile arrays, a general technique for the underlying inference still remains an open question~\cite{luo2017robotic}. Visual and depth-based tracking have been widely studied~\cite{schmidt2015dart}, but suffer from occlusion due to clutter or self-occlusions with the gripper or robot. We provide a general formulation of pure tactile inference, that could later accommodate additional sensing modalities. 

Pure tactile inference is challenging because, unlike vision, touch cannot directly provide global estimates of object model or pose. Instead, it provides detailed, local information that must be fused into a global model. Moreover, touch is intrusive: the act of sensing itself constantly perturbs the object. We consider tactile inference as an analog of the well-studied simultaneous localization and mapping (SLAM) problem in mobile robotics~\cite{durrant2006simultaneous}. Errors in object tracking accumulate to affect its predicted shape, and vice versa.
\begin{figure}[t]
	\centering
	\includegraphics[width=0.8\columnwidth]{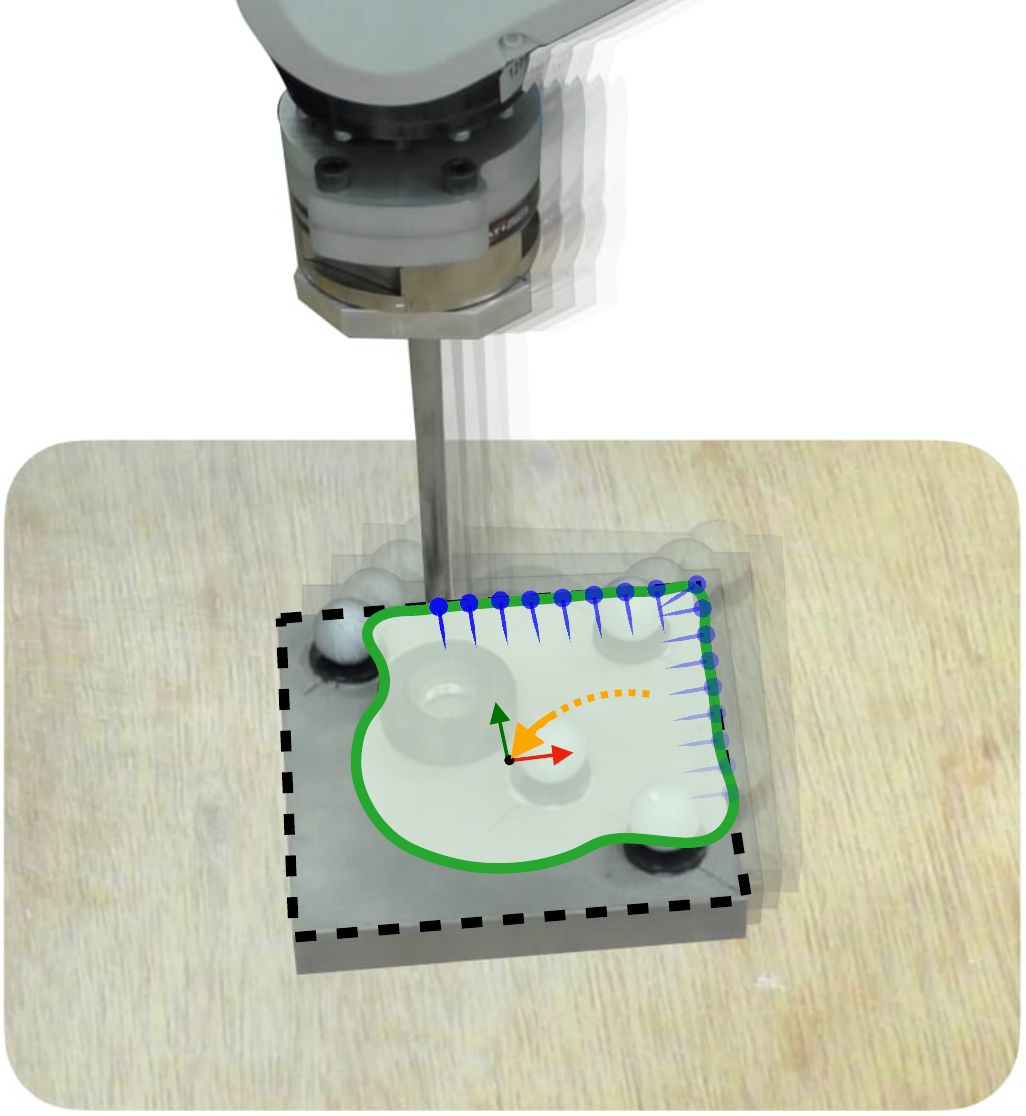}
	\caption{The shape and pose estimation problem in a pusher-slider system. A robot manipulator pushes along a planar object, while recording a stream of tactile measurements. Our method builds a shape contour in real-time as a Gaussian process implicit surface, and optimizes for pose via geometry and physics-based constraints. Figure shows tactile measurements (\protect\tikz[baseline=-0.6ex]\protect\draw[contact_color,fill=contact_color, line width=0.5pt] (0,0) circle (.3ex);), estimated motion (\textcolor{motion_color}{$\mathbf{\downarrow}$}), estimated shape/pose (\protect\tikz[baseline=-0.6ex]\protect\draw[shape_color,fill=shape_fill_color, line width=1.0pt] (0,0) circle (.7ex);), and ground-truth (\textbf{-\,-}).}
	\label{fig:cover} 
	\GobbleLarge
\end{figure}

Central to this problem is choosing a shape representation that both faithfully approximates arbitrary geometries, and is amenable to probabilistic updates. This excludes most parametric models such as polygons/polyhedrons~\cite{yu2015shape}, superquadrics~\cite{solina1990recovery}, voxel maps~\cite{duncan2013multi}, point-clouds~\cite{meier2011probabilistic}, and standard meshes~\cite{varley2017shape}. Gaussian process implicit surfaces (GPIS)~\cite{williams2007gaussian} are one such nonparametric shape representation that satisfies these requirements. 

In this paper, we demonstrate online shape and pose estimation for a planar pusher-slider system. We perform tactile exploration of the object via contour following, that generates a stream of contact and force measurements. Our novel schema combines efficient GPIS regression with factor graph optimization over geometric and physics-based constraints. The problem is depicted in Fig. \ref{fig:cover}: the pusher moves along the object while estimating its shape and pose. 

We expand the scope of the batch-SLAM method by Yu \etal~\cite{yu2015shape} with a more meaningful shape representation, and real-time online inference. Our contributions are:
\begin{enumerate}
\item[{(1)}] A formulation of the tactile SLAM problem that alternates between GPIS shape regression and sparse nonlinear incremental pose optimization,
\item[{(2)}] Efficient implicit surface generation from touch using overlapping local Gaussian processes (GPs),  
\item[{(3)}] Fixed-lag smoothing over contact, geometry, and frictional pushing mechanics to localize objects,  
\item[{(4)}] Results from tactile exploration across different planar object shapes in simulated and real settings.
\end{enumerate}
%

\section{Related work}
\label{sec:related}

\subsection{SLAM and object manipulation}
\label{ssec:related_1}

Our work is closely related to that of Yu \etal~\cite{yu2015shape}, that recovers shape and pose from tactile exploration of a planar object. The approach uses contact measurements and the well-understood mechanics of planar pushing~\cite{lynch1992manipulation, goyal1991planar} as constraints in a batch optimization. Naturally, this is expensive and unsuitable for online tactile inference. Moreover, the object shape is represented as ordered control points, to form a piecewise-linear polygonal approximation. Such a representation poorly approximates arbitrary objects, and fails when data-association is incorrect. Moll and Erdmann~\cite{moll2004reconstructing} consider the illustrative case of reconstructing motion and shape of smooth, convex objects between two planar palms. Strub \etal~\cite{strub2014correcting} demonstrate the full SLAM problem with a dexterous hand equipped with tactile arrays. 

Contemporaneous research considers one of two simplifying assumptions: modeling shape with fixed pose~\cite{meier2011probabilistic, dragiev2011gaussian, martinez2013active, yi2016active, driess2019active}, or localizing with known shape~\cite{petrovskaya2011global, zhang2013dynamic, koval2015pose, bauza2019tactile, Sodhi21icra}. The extended Kalman filter (EKF) has been used in visuo-tactile methods~\cite{hebert2011fusion, izatt2017tracking}, but is prone to linearization errors. At each timestep, it linearizes about a potentially incorrect current estimate, leading to inaccurate results. Smoothing methods~\cite{kaess2012isam2} are more accurate as they preserve a temporal history of costs, and solve a nonlinear least-squares problem. These frameworks have been used to track known objects with vision and touch~\cite{yu2018realtime, lambert2019joint}, and rich tactile sensors~\cite{, Sodhi21icra}.

\subsection{Gaussian process implicit surfaces}
\label{ssec:related_2}

A continuous model which can generalize without discretization errors is of interest to global shape perception. Implicit surfaces have long been used for their smoothness and ability to express arbitrary topologies~\cite{blinn1982generalization}. Using Gaussian processes~\cite{rasmussen2003gaussian} as their surface potential enables probabilistic fusion of  noisy measurements, and reasoning about shape uncertainty. GPIS were formalized by Williams and Fitzgibbon~\cite{williams2007gaussian}, and were later used by Dragiev \etal~to learn shape from grasping~\cite{dragiev2011gaussian}. It represents objects as a signed distance field (SDF): the signed distance of spatial grid points to the nearest object surface. The SDF and surface uncertainty were subsequently used for active tactile exploration~\cite{dragiev2013uncertainty, li2016dexterous, yi2016active}. To our knowledge, no methods use GPIS alongside pose estimation for manipulation tasks.

Online GP regression scales poorly due to the growing cost of matrix inversion~\cite{rasmussen2003gaussian}. Spatial mapping applications address this by either sparse approximations to the full GP~\cite{snelson2006sparse}, or training separate local GPs~\cite{kim2013continuous, stork2020ensemble}. Lee \etal~\cite{lee2019online} propose efficient, incremental updates to the GPIS map through spatial partitioning.
\begin{figure}[t]
	\centering
	\includegraphics[width=\columnwidth]{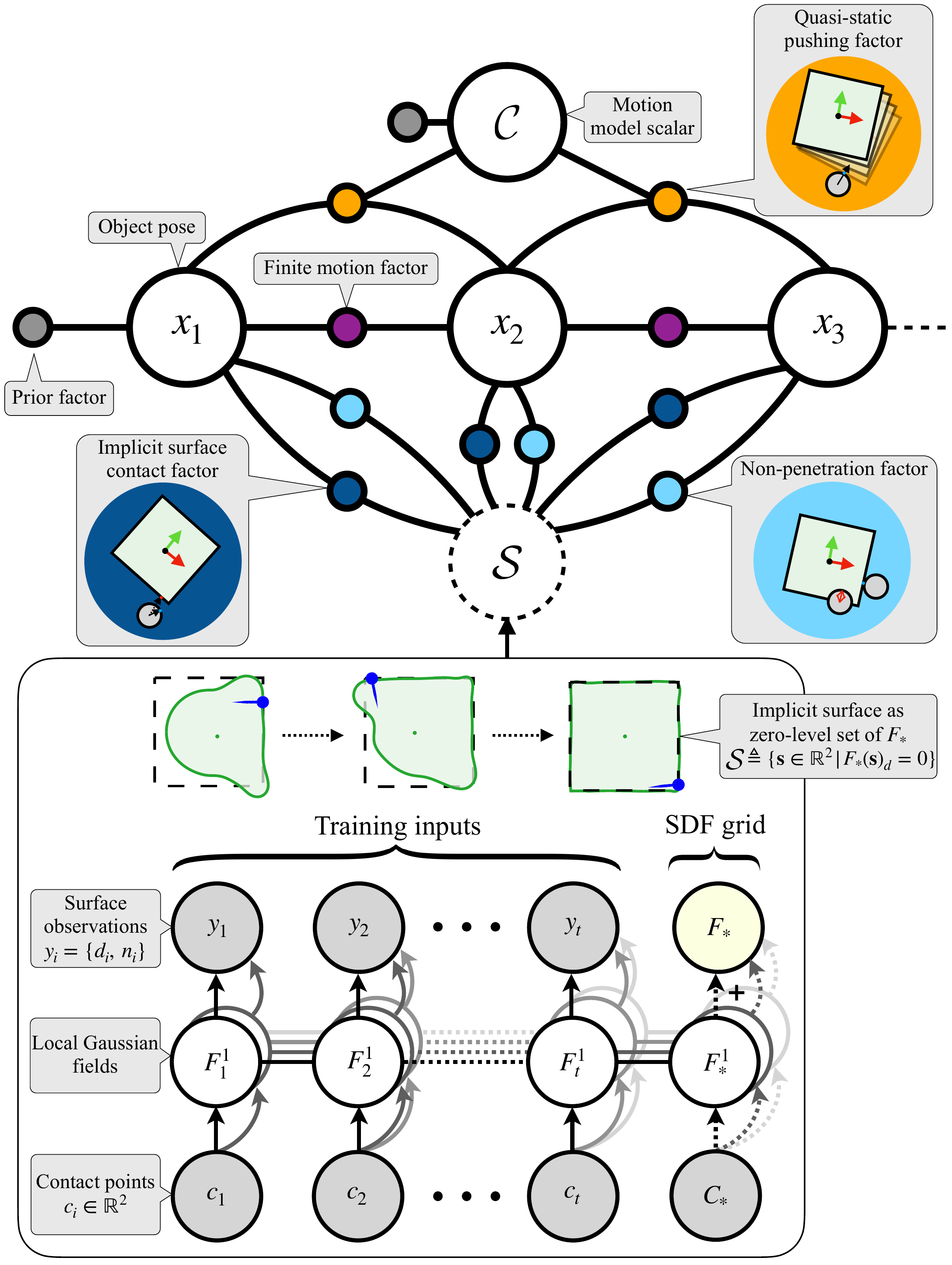}
	\caption{The combined formulation between the factor graph (Section \ref{sec:fac_graph}) and GPIS (Section \ref{sec:gpis}). \textbf{[top]} The graph illustrates the relationship between the variables to be optimized for (circles) and the factors that act as constraints (colored dots). \textbf{[bottom]} Our GPIS builds an implicit surface shape representation that is the zero level-set of GP potential function. Spatial partitioning with local GPs enables efficient regression.}
	\label{fig:final_graph} 
	\GobbleLarge
\end{figure}
%

\section{Problem formulation: tactile slam}
\label{sec:problem}

We consider a rigid planar object on a frictional surface, moved around by a pusher (see Fig. \ref{fig:cover}). The interaction is governed by simple contour following for tactile exploration. Given a stream of tactile measurements, we estimate the 2-D shape and object's planar pose in real-time. 

\boldsubheading{Object pose}
The object pose at the current timestep $t$ is defined by position and orientation $\mathbf{x}_t = (x, \ y, \ \theta) \in SE\left(2\right)$. 

\boldsubheading{Object shape}
The estimated object shape is represented as an implicit surface $\mathbf{\mathcal{S}} \in \mathbb{R}^2$ in the reference frame of $\mathbf{x}_t$.

\boldsubheading{Tactile measurements}
At every timestep we observe:
\begin{equation}
\mathbf{z}_t = \Big\{ \underbrace{\mathbf{p}_t \in \mathbb{R}^{2}}_{\text{pusher position}}, \ \underbrace{\mathbf{f}_t \in \mathbb{R}^{2}}_{\text{force vector}}, \ \underbrace{{\Theta}_t \in \{0, 1\}}_{\text{contact/no-contact}}  \Big\}
\label{eq:1}
\end{equation}
Pusher position is obtained from the robot's distal joint state, and force is sensed directly from a force/torque sensor. Contact is detected with a minimum force threshold, and the estimated contact point $\mathbf{c}_t$  is derived from knowledge of $\mathbf{p}_t$, $\mathbf{f}_t$ and probe radius $r_{\text{probe}}$. This is consistent with the formulation in \cite{yu2015shape}, with the addition of direct force measurements $\mathbf{f}_t$. For simplicity we consider a single pusher, but it can be expanded to multiple pushers or a tactile array. 

\boldsubheading{Assumptions}
We make a minimal set of assumptions, similar to prior work in planar pushing~\cite{yu2015shape, yu2018realtime, lambert2019joint}: 
\begin{itemize}[leftmargin=1em]
    \item Quasi-static interactions and limit surface model~\cite{goyal1991planar, lee1991fixture},
    \item Uniform friction and pressure distribution between bodies,
    \item Object's rough scale and initial pose, and 
    \item No out-of-plane effects.
\end{itemize}
%
 
The remainder of the paper is organized as such: Section \ref{sec:gpis} formulates building shape estimate $\mathbf{\mathcal{S}}$ as the implicit surface of a GP, given current pose estimate and measurement stream. Section \ref{sec:fac_graph} describes factor graph inference to obtain pose estimate $\mathbf{x}_t^*$ with $\mathbf{\mathcal{S}}$ and the measurement stream. These two processes are carried out back-and-forth for online shape and pose hypothesis at each timestep (Fig. \ref{fig:final_graph}). Finally, we demonstrate its use in simulated and real experiments (Section \ref{sec:expts}) and present concluding remarks (Section \ref{sec:conc}). 

\section{Shape estimation with implicit surfaces}
\label{sec:gpis}

\subsection{Gaussian process implicit surface}
\label{ssec:gpis_1}
A Gaussian process learns a continuous, nonlinear function from sparse, noisy datapoints~\cite{rasmussen2003gaussian}. Surface measurements are in the form of contact points $\mathbf{c}_i$ and normals $\mathbf{n}_{i}$, transformed to an object-centric reference frame. Given $N$ datapoints, the GP learns a mapping $X \mapsto Y$ between them:
\begin{equation}
F: \underbrace{\{ c_{i_x}, \ c_{i_y} \}^{i = 1 \cdots N}}_{X \in \, \mathbb{R}^2} \mathbf{ \ \mapsto \ }  \underbrace{\{d_i, n_{i_x}, n_{i_y}\}^{i = 1 \cdots N}}_{Y  \in  \, \mathbb{R}^3}
\label{eq:2}
\end{equation}
\begin{equation}
\parbox{10em}{\centering \footnotesize where $d$ represents \\ signed-distance from \\ object surface} \ {\footnotesize \begin{cases}
      d = 0,&\text{on surface} \\
      d < 0,&\text{inside object} \\
      d > 0,&\text{outside object}
    \end{cases}}
\label{eq:3}
\end{equation}

The posterior distribution at a test point $\mathbf{c}_*$ is shown to be $F_* \sim \mathcal{GP}(\bar{F_*}, \sigma_*^2)$, with output mean and variance~\cite{rasmussen2003gaussian}: 
\begin{equation}
\begin{split}
\bar{F_*} &= k_*^T\left(K + \sigma_{\text{noise}}^2I \right)^{-1}Y \\
\sigma_*^2 &= k_{**} - k_*^T\left(K + \sigma_{\text{noise}}^2I \right)^{-1}k_*
\end{split}
\label{eq:4}
\end{equation}
where $K \in \mathbb{R}^{N \! \times \! N}$, $k_* \in \mathbb{R}^{N \! \times \! 1}$ and $k_{**} \in \mathbb{R}$ are the train-train, train-test, and test-test kernels respectively. We use a thin-plate kernel~\cite{williams2007gaussian}, with hyperparameter tuned for scene dimensions. The noise in output space is defined by a zero-mean Gaussian with variance $\sigma_{\text{noise}}^2$. While contact points condition the GP on zero SDF observations, contact normals provide function gradient observations~\cite{dragiev2011gaussian}. Thus, we can jointly model both SDF and surface direction for objects.

We sample the posterior over an $M$ element spatial grid of test points $\mathbf{C}_*$, to get SDF $F_{*_{d}}$. The estimated implicit surface $\mathcal{S}$ is then the zero-level set contour:
\begin{equation}
\mathcal{S} \triangleq \{ \mathbf{s} \in \mathbb{R}^2 \ | \  F_*(\mathbf{s})_d = 0 \} 
\label{eq:5}
\end{equation}
The zero-level set $\mathcal{S}$ is obtained through a contouring subroutine on $F_{*_{d}}$. $\mathcal{S}$ is initialized with a circular prior and updated with every new measurement $\{\mathbf{c}_i, \mathbf{n}_{i}, d \! = \! 0 \}$. In Fig. \ref{fig:gpis_butter} we reconstruct the \texttt{butter} shape~\cite{yu2016more} with a sequence of noisy contact measurements. 

\subsection{Efficient online GPIS}
\label{ssec:gpis_2}
\begin{figure}[t]
	\centering
	\includegraphics[width=\columnwidth]{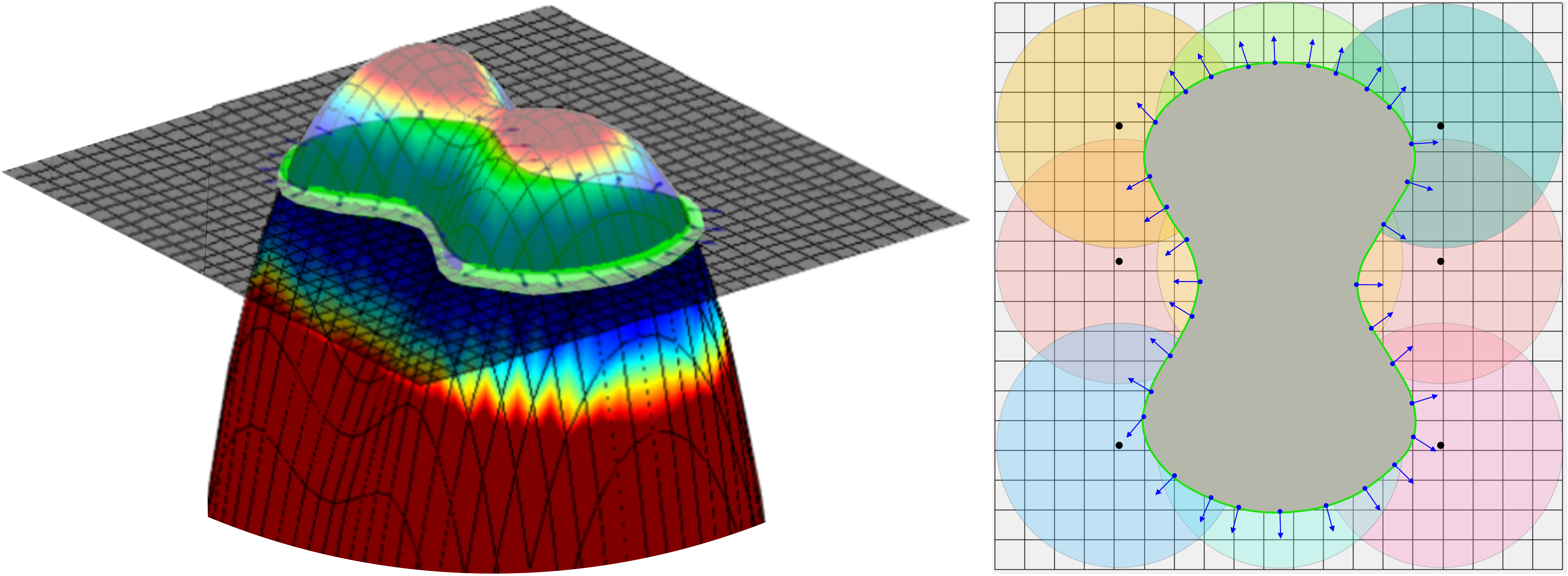}
	\caption{\textbf{[left]} Gaussian process potential function and its implicit surface (green) for noisy contact measurements on the \texttt{butter} shape~\cite{yu2016more}. The colormap shows spatial grid uncertainty. \textbf{[right]} The overlapping local GPs $F^1 \ldots F^L$. Each GP is responsible for training and test points within its radius, and the overlapping regions ensure continuity in the shape contour.}
	\label{fig:gpis_butter} 
	\GobbleLarge
\end{figure}

%
In a na\"ive GP implementation, the computational cost restricts online shape perception. Equation \ref{eq:4} requires an $N \! \times \! N$ matrix inversion that is $O(N^3)$, and spatial grid testing that is $O(MN^2)$. We use local GPs and a subset of training data approximation for efficient online regression: 

\boldsubheading{Local GP regression} 
We adopt a spatial partitioning approach similar to~\cite{kim2013continuous, lee2019online, stork2020ensemble}. The scene is divided into $L$ independent local GPs $F^1 \ldots F^L$, each with a radius $r$ and origin (Fig. \ref{fig:gpis_butter}). Each $F^i$ claims the training and test points that fall within $r$ of its origin. The GPs effectively govern smaller domains ($N_{F^i} \! \ll \! N$ and $M_{F^i} \! \ll \! M$), and not the entirety of the spatial grid. At every timestep: (i) a subset of local GPs are updated,  (ii) only the relevant test points are resampled. Kim \etal~\cite{kim2013continuous} demonstrate the need for overlapping domains to avoid contour discontinuity at the boundaries. Thus, we increase $r$,  and in overlapping regions, the predicted estimates are averaged among GPs. 

\boldsubheading{Subset of data} 
Before adding a training point $\mathbf{c}_i$ to the active set, we ensure that the output variance $\sigma_i^2$ is greater than a pre-defined threshold $\sigma_{\text{thresh}}^2$. This promotes sparsity in our model by excluding potentially redundant information.
%

\boldsubheading{Implementation} 
Rather than direct matrix inversions (Equation \ref{eq:4}), it is more efficient to use the Cholesky factor of the kernel matrix~\cite{rasmussen2003gaussian}. In our online setting, we directly update the Cholesky factor $\mathcal{L} \mathcal{L}^T = \left(K + \sigma_{\text{noise}}^2I \right)$ with new information. We multi-thread the update/test steps, and perform these at a lower frequency than the graph optimization. We set $L = 25$, grid sampling resolution $5$~mm, and circular prior of radius $40$~mm for our experiments.

\section{Pose estimation with factor graphs}
\label{sec:fac_graph}
\subsection{Factor graph formulation}
\label{ssec:fac_graph_1}
The \textit{maximum a posteriori} (MAP) estimation problem gives variables that maximally agree with the sensor measurements. This is commonly depicted as a factor graph: a bipartite graph with variables to be optimized for and factors that act as constraints (Fig. \ref{fig:final_graph}). The augmented state comprises object poses and a motion model scalar $\mathcal{C}$: 
\begin{equation}
\mathbf{\mathcal{X}}_t = \{ \mathbf{x}_{1}, \ldots, \mathbf{x}_t \ ; \ \mathbf{\mathcal{C}} \}
\label{eq:6}
\end{equation}
Prior work in pushing empirically validates measurement noise to be well-approximated by a Gaussian distribution~\cite{yu2018realtime}. With Gaussian noise models, MAP estimation reduces to a nonlinear least-squares problem~\cite{dellaert2017factor}. Our MAP solution (given best-estimate shape $\mathcal{S}$) is: 
\begin{equation}
\footnotesize
\begin{split}
\label{eq:7}
&\mathcal{X}^*_t =  \underset{\mathcal{X}_t}{\operatorname{argmin}} \sum_{i=1}^t  \Big( \underbrace{ \mystrut{1.5ex} ||Q(\mathbf{x}_{i-1},\mathbf{x}_{i}, \mathbf{z}_{i - 1}, \mathcal{C})||_{\Sigma_{Q}}^2}_{\text{\tikz[baseline=-0.6ex]\draw[black,fill=qcolor, line width=0.5pt] (0,0) circle (.5ex);~QS pushing factor}}
+
\underbrace{ \mystrut{1.5ex} ||I(\mathbf{x}_{i}, \mathbf{z}_{i}, \mathcal{S})||_{\Sigma_{I}}^2}_{\text{\tikz[baseline=-0.6ex]\draw[black,fill=icolor, line width=0.5pt] (0,0) circle (.5ex);~IS contact factor}}\\
&+
\underbrace{ \mystrut{1.5ex} ||P(\mathbf{x}_{i}, \mathbf{z}_{i}, \mathcal{S})||_{\Sigma_{P}}^2}_{\text{\tikz[baseline=-0.6ex]\draw[black,fill=pcolor, line width=0.5pt] (0,0) circle (.5ex);~Non-penetration factor}}
+ \underbrace{ \mystrut{1.5ex} ||F(\mathbf{x}_{i-1},\mathbf{x}_{i})||_{\Sigma_{F}}^2}_{\text{\tikz[baseline=-0.6ex]\draw[black,fill=fcolor, line width=0.5pt] (0,0) circle (.5ex);~Finite motion factor}} \Big)
+ \underbrace{ \mystrut{1.5ex} ||\mathbf{p}_0||_{\Sigma_0}^2 + ||\mathbf{c}_0||_{\Sigma_{c}}^2}_{\text{\tikz[baseline=-0.6ex]\draw[black,fill=prcolor, line width=0.5pt] (0,0) circle (.5ex);~Priors}} 
\end{split}
\normalsize
\raisetag{35pt}
\end{equation}

This is graphically represented in Fig. \ref{fig:final_graph}, and the cost functions are described in Section \ref{ssec:fac_graph_2}. Given covariance matrix $\Sigma$, $\left\Vert v\right\Vert_{\Sigma}^{2}=v^{T}\Sigma^{-1}v$ is the Mahalanobis distance of $v$. The noise terms for covariances \{$\Sigma_{Q}, \ldots, \Sigma_{c}$\} are empirically selected. The online estimation is performed using incremental smoothing and mapping (iSAM2)~\cite{kaess2012isam2}. Rather than re-calculating the entire system every timestep, iSAM2 updates the previous matrix factorization with new measurements. In addition, we use a fixed-lag smoother to bound optimization time over the exploration~\cite{dellaert2017factor}. Fixed-lag smoothing maintains a fixed temporal window of states $\mathcal{X}_w$, while efficiently marginalizing out preceding states ($100$ timesteps in our experiments). Note that this is different from simply culling old states and discarding information. 

\subsection{Cost functions}
\label{ssec:fac_graph_2}
\begin{figure}[b]
	\centering
	\includegraphics[width=\columnwidth]{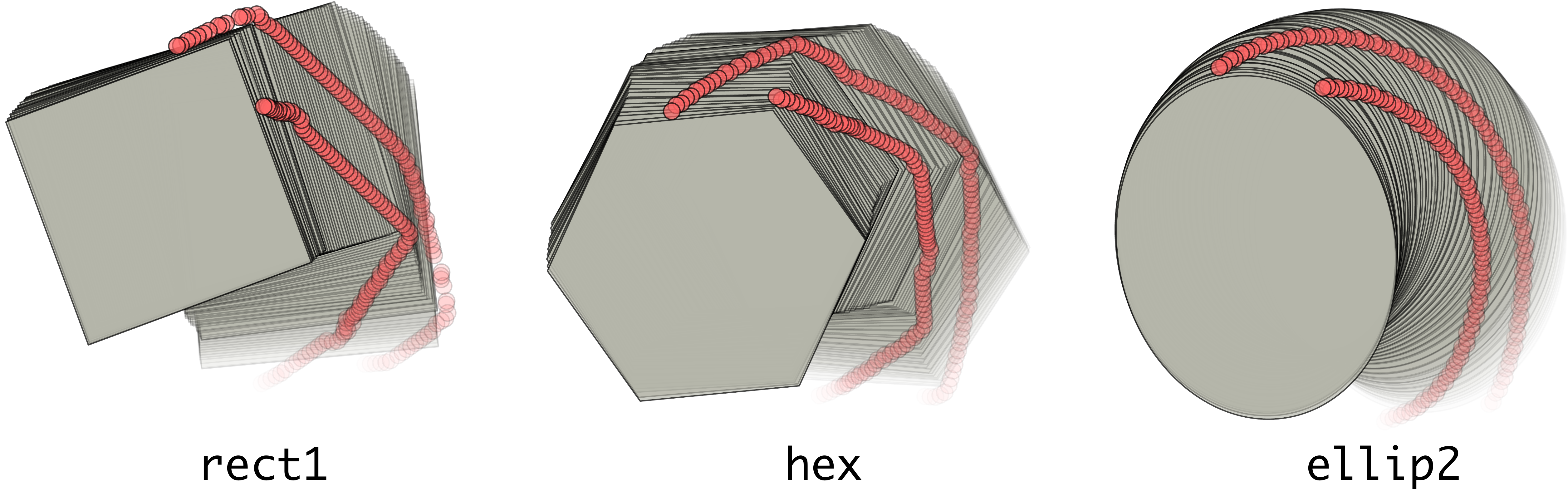}
	\caption{Snippets of the tactile exploration data collected in the PyBullet simulator. We use a two-finger pusher to perform contour following, and collect the tactile measurements and ground-truth poses.}
	\label{fig:sim_data} 
\end{figure}
\boldsubheading{\tikz[baseline=-0.6ex]\draw[black,fill=qcolor, line width=1pt] (0,0) circle (.5ex);~QS pushing factor}
The quasi-static model uses a limit surface (LS) model to map between pusher force $\textbf{f}_t$ and object motion~\cite{goyal1991planar}. Specifically, Lynch \etal~\cite{lynch1992manipulation} develop an analytical model using an ellipsoid LS approximation~\cite{lee1991fixture}. The factor ensures object pose transitions obey the quasi-static motion model, with an error term: 
\begin{equation}
Q(\mathbf{x}_{t-1},\mathbf{x}_{t}, \mathbf{z}_{t - 1}, \mathcal{C}) = \bigg[ \frac{v_x}{\omega} - \mathcal{C}^2 \frac{{f_{t - 1}}_x}{\tau},\frac{v_y}{\omega} - \mathcal{C}^2 \frac{{f_{t - 1}}_y}{\tau} \bigg]
\label{eq:8}
\end{equation}
\begin{itemize}
 \item $(v_x, v_y, \omega)$ is the object's velocity between $\mathbf{x}_{t-1}$ and $\mathbf{x}_{t}$,
 \item $\tau$ is the applied moment w.r.t. pose center of $\mathbf{x}_{t-1}$,
 \item $\mathcal{C} = {\tau_{\text{max}}}/{f_{\text{max}}}$ is an object-specific scalar ratio dependent on pressure distribution. 
\end{itemize}  
For a more rigorous treatment, we refer the reader to~\cite{yu2015shape}. We weakly initialize $\mathcal{C}$ with our known circular shape prior, and incorporate it into the optimization. 

\boldsubheading{\tikz[baseline=-0.6ex]\draw[black,fill=icolor, line width=1pt] (0,0) circle (.5ex);~Implicit surface contact factor}
Given ${\Theta}_t \! = \! 1$ (contact), we encourage the measured contact point $\textbf{c}_t$ to lie on the manifold of the object. We define $\mathbf{\Phi} = \mathbf{\Phi}(\mathbf{x}_t, \mathbf{z}_t, \mathcal{S})$ that computes the closest point on $\mathcal{S}$ w.r.t. the pusher via projection. The error term is defined as: 
\begin{equation}
I(\mathbf{x}_{t}, \mathbf{z}_{t}, \mathcal{S}) = \big[ \mathbf{\Phi} - \mathbf{c}_t \ , \ \| \mathbf{\Phi} - \mathbf{p}_t \| - r_{\text{probe}} \big]
\label{eq:9}
\end{equation}
This ensures that in the presence of noise, the contact points lie on the surface and the normals are physically valid \cite{yu2015shape}. 

\boldsubheading{\tikz[baseline=-0.6ex]\draw[black,fill=pcolor, line width=1pt] (0,0) circle (.5ex);~Non-penetration factor}
While we assume persistent contact, this cannot be assured when there is more than one pusher. When ${\Theta}_t \! = \! 0$ (no contact) we enforce an intersection penalty (as used in \cite{lambert2019joint}) on the pusher-slider system. We define $\mathbf{\Psi} = \mathbf{\Psi}(\mathbf{x}_t, \mathbf{z}_t, \mathcal{S})$ to estimate the pusher point furthest inside the implicit surface, if intersecting. The error is:
\begin{align}
    P(\mathbf{x}_{t}, \mathbf{z}_{t}, \mathcal{S}) = 
    \begin{cases}
    \| \mathbf{\Psi} - \mathbf{\Phi} \|, &\text{when intersecting}  \\
    0, &\text{when not intersecting}
    \end{cases}
    \label{eq:10}
\end{align}

\boldsubheading{\tikz[baseline=-0.6ex]\draw[black,fill=fcolor, line width=1pt] (0,0) circle (.5ex);~Finite motion factor}
Given persistent contact, we weakly bias the object towards constant motion in $SE\left(2\right)$. The magnitude is empirically chosen from the planar pushing experiments. We observe this both smooths the trajectories, and prevents an indeterminate optimization.

\boldsubheading{\tikz[baseline=-0.6ex]\draw[black,fill=prcolor, line width=1pt] (0,0) circle (.5ex);~Priors}
The prior $\mathbf{p}_0$ anchors the optimization to the initial pose. $\mathcal{C}$ is initialized with $\mathbf{c}_0$ using the circular shape prior. 

\section{Experimental evaluation}
\label{sec:expts}
\begin{figure}[b]
    \GobbleMedium
	\centering
	\includegraphics[width=\columnwidth]{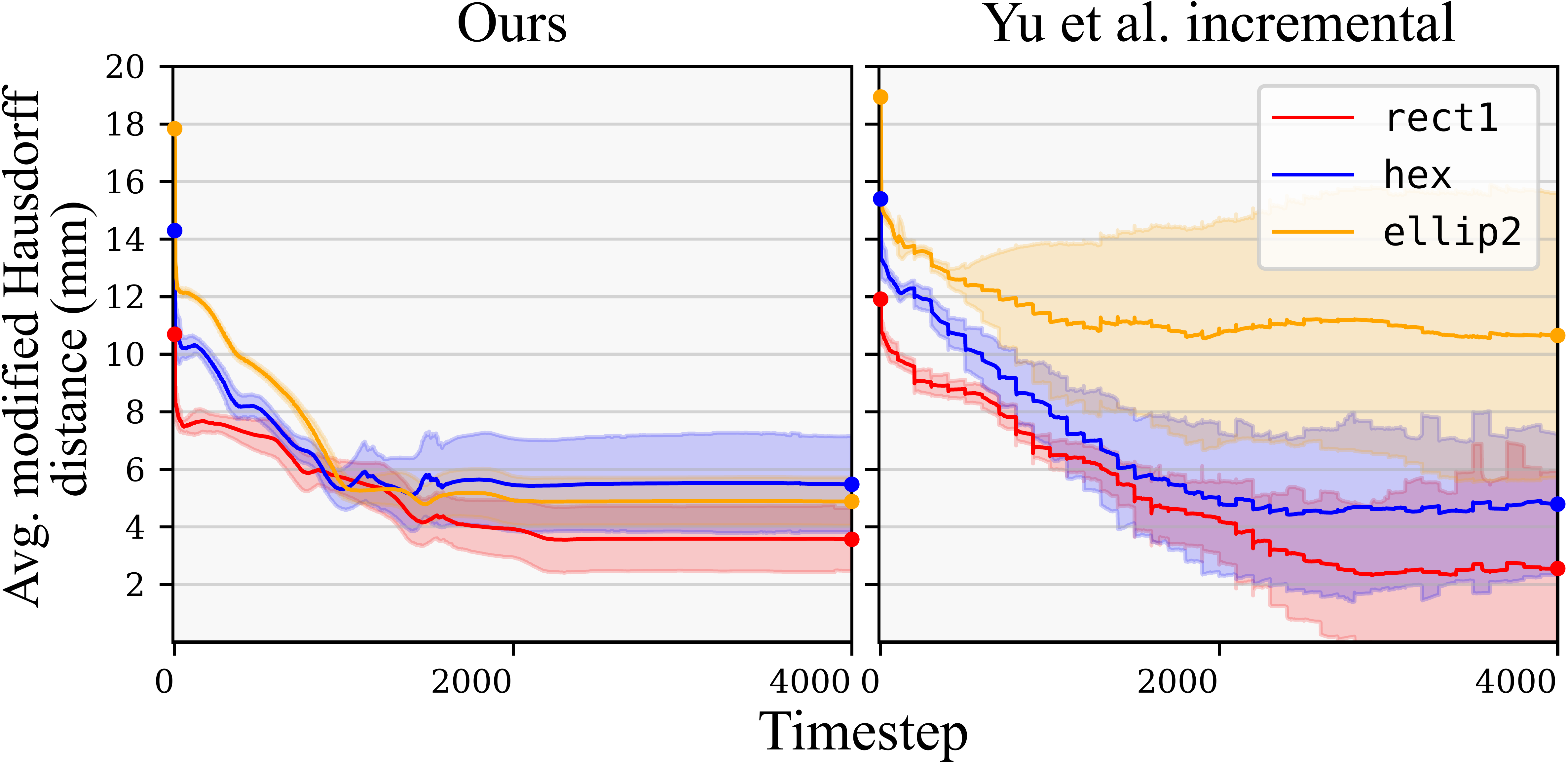}
	\caption{Average MHD w.r.t. the ground-truth model for the $50$ logs and for each of the $3$ objects. Comparing ours \textbf{[left]} to \textit{Yu \etal~incremental} \textbf{[right]}, we see less variance across all shapes, and much lower error in \texttt{ellip2}. This shows that while the piecewise-linear representation is suited to polygonal objects, it understandably fails to generalize to more arbitrary shapes. The GPIS faithfully approximates both classes. Moreover, the errors in data association lead to large variance among trials in \texttt{rect1} and \texttt{hex}.}
	\label{fig:sim_MHD} 
\end{figure}
\begin{figure*}
\centering
\begin{minipage}[b]{.59\textwidth}
\includegraphics[width=\textwidth]{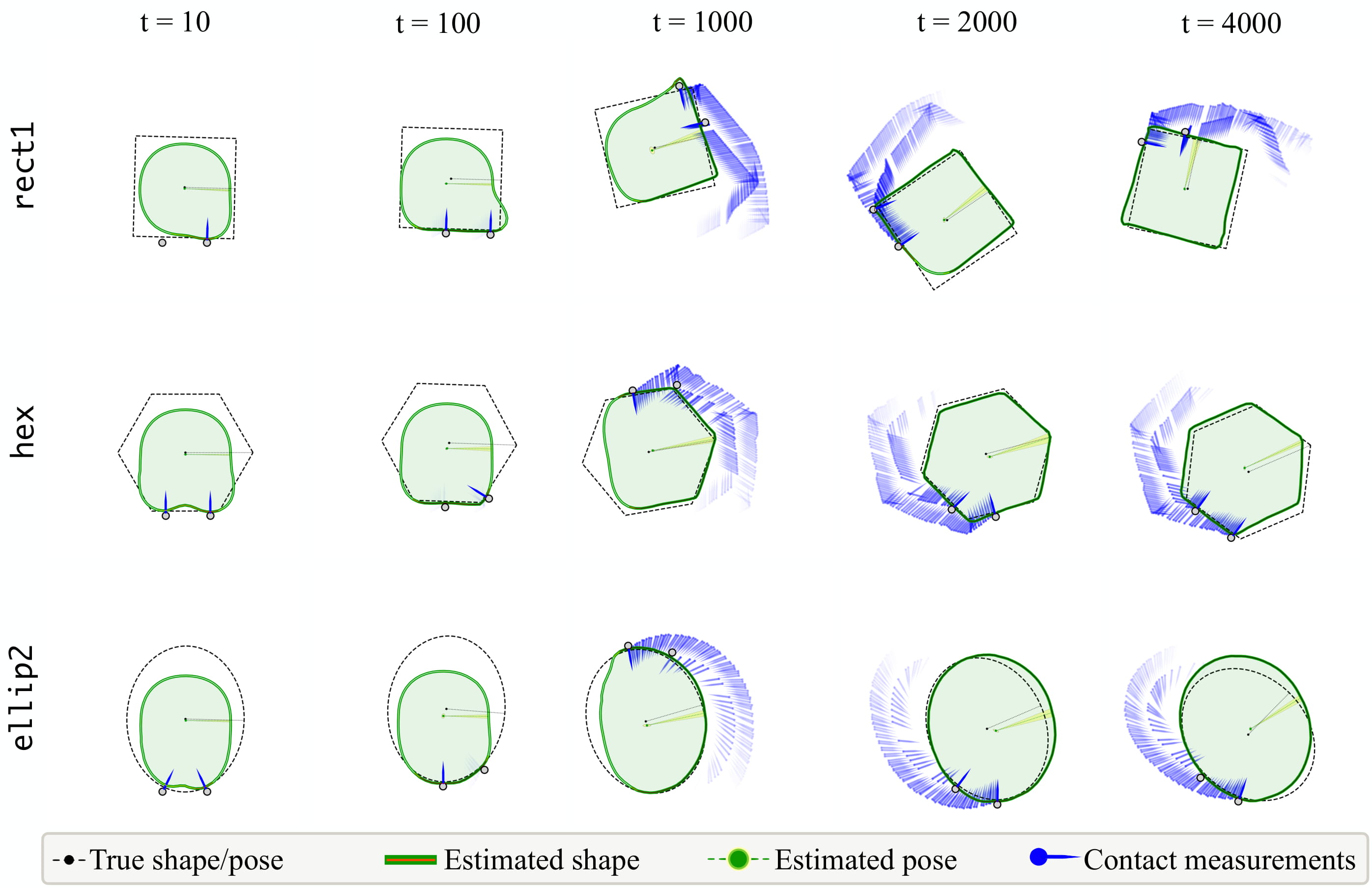}
\caption{Estimated shape and pose of representative simulation trials, with timesteps $t$. We compare these against the ground-truth, and overlay the stream of tactile measurements.}\label{fig:sim_pose_shape}
\end{minipage}\qquad
\begin{minipage}[b]{.32\textwidth}
\includegraphics[width=\textwidth]{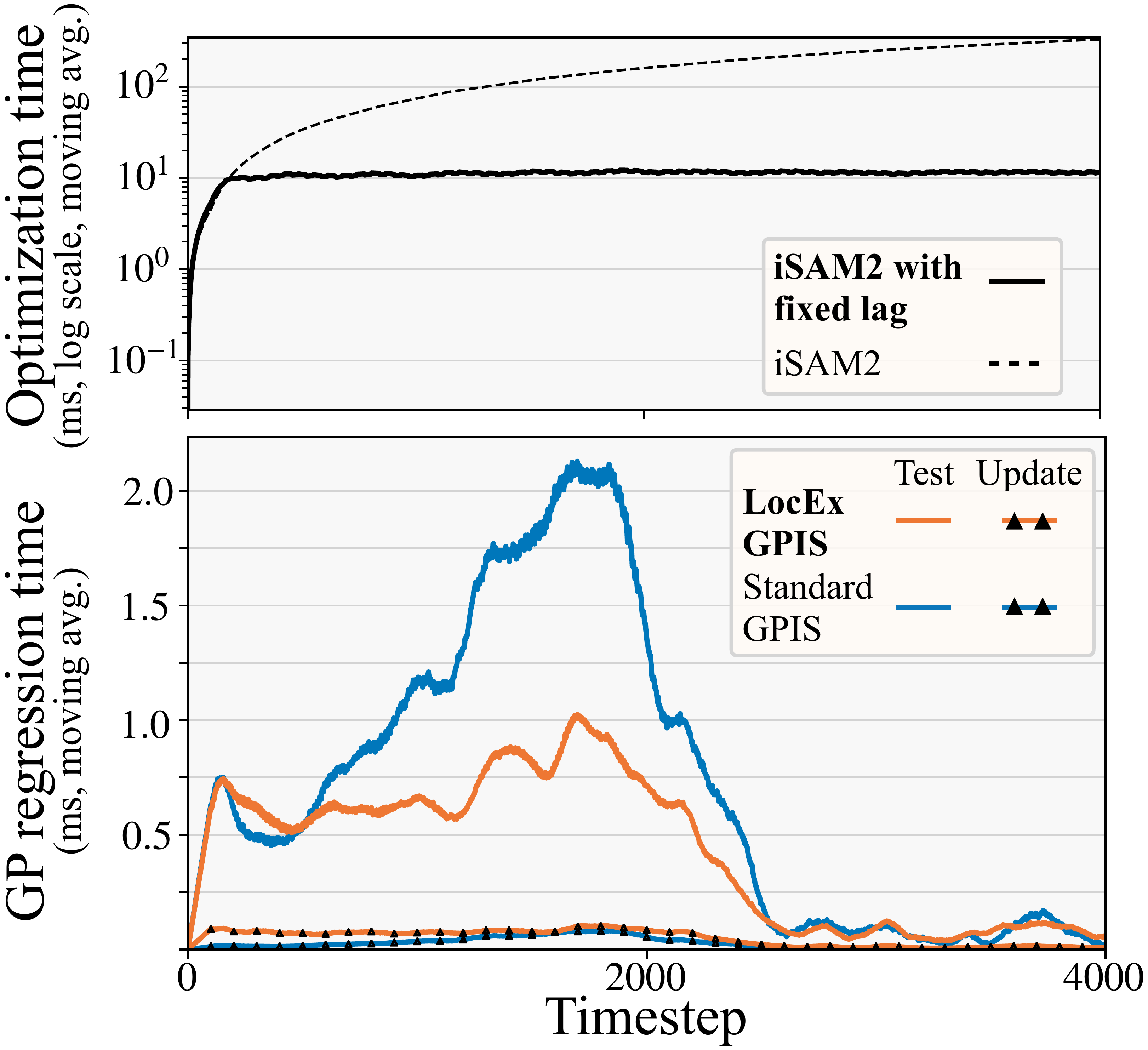}
\caption{Moving average of execution time of the processes, over all $3 \! \times \! 50$ logs. \textbf{[top]} While the complexity of full iSAM2 grows linearly with time, fixed-lag smoothing maintains a bounded optimization window. \textbf{[bottom]} As a result of spatial partitioning, local GPIS regression has a lower query time compared to the standard implementation.}\label{fig:timing_plot}
\end{minipage}
\GobbleLarge
\end{figure*}
We demonstrate the framework in both simulated (Section \ref{ssec:expts_1}) and real-world planar pushing tasks (Section \ref{ssec:expts_2}).

\boldsubheading{Evaluation metrics}
For pose error, we evaluate the root mean squared error (RMSE) in translation and rotation w.r.t. the true object pose. For shape, we use the modified Hausdorff distance (MHD)~\cite{dubuisson1994modified} w.r.t. the true object model. The Hausdorff distance is a measure of similarity between arbitrary shapes that has been used to benchmark GPIS mapping~\cite{stork2020ensemble}, and the MHD is an improved metric more robust to outliers. 

\boldsubheading{Baseline}
We compare against the work of Yu \etal~\cite{yu2015shape}, which recovers shape and pose as a batch optimization (Section \ref{ssec:related_1}). For fairness, we implement an online version as our baseline, that we refer to as \textit{Yu \etal~incremental}. We use $N_s = 25$ shape nodes in the optimization to better represent non-polygonal objects.  

\boldsubheading{Compute}
We use the GTSAM library with iSAM2~\cite{kaess2012isam2} for incremental factor graph optimization. The experiments were carried out on an Intel Core i7-7820HQ CPU, 32GB RAM without GPU parallelization.
%
\subsection{Simulation experiments}
\label{ssec:expts_1}
\boldsubheading{Setup}
The simulation experiments are conducted in PyBullet~\cite{coumans2010bullet} (Fig. \ref{fig:sim_data}). We use a two-finger pusher ($r_{\text{probe}} = 6.25$~mm) to perform tactile exploration at $60$~mm/s. Contour following is based on previous position and sensed normal reaction. The coefficients of friction of the object-pusher and object-surface are both $0.25$. Zero-mean Gaussian noise with std. dev. $\left[0.1~\text{mm},~0.01~\text{N}\right]$ are added to $\left[\textbf{c}_t,~\textbf{f}_t\right]$.

We run $50$ trials of $4000$ timesteps each, on three shape models~\cite{yu2016more}: \textbf{(i)} \texttt{rect1} ($90$~mm side), \textbf{(ii)} \texttt{hex} ($60.5$~mm circumradius), and \textbf{(iii)} \texttt{ellip2} ($130.9$~mm maj. axis). While our framework can infer arbitrary geometries, the contour following schema is best suited for simpler planar objects. The object's initial pose $\mathbf{x}_0$ is randomly perturbed over the range of $\pm (2~\text{mm}, \! \ 2~\text{mm}, \! \ 5^{\circ})$.
\begin{figure}[!b]
	\GobbleMedium
	\centering
	\includegraphics[width=\columnwidth]{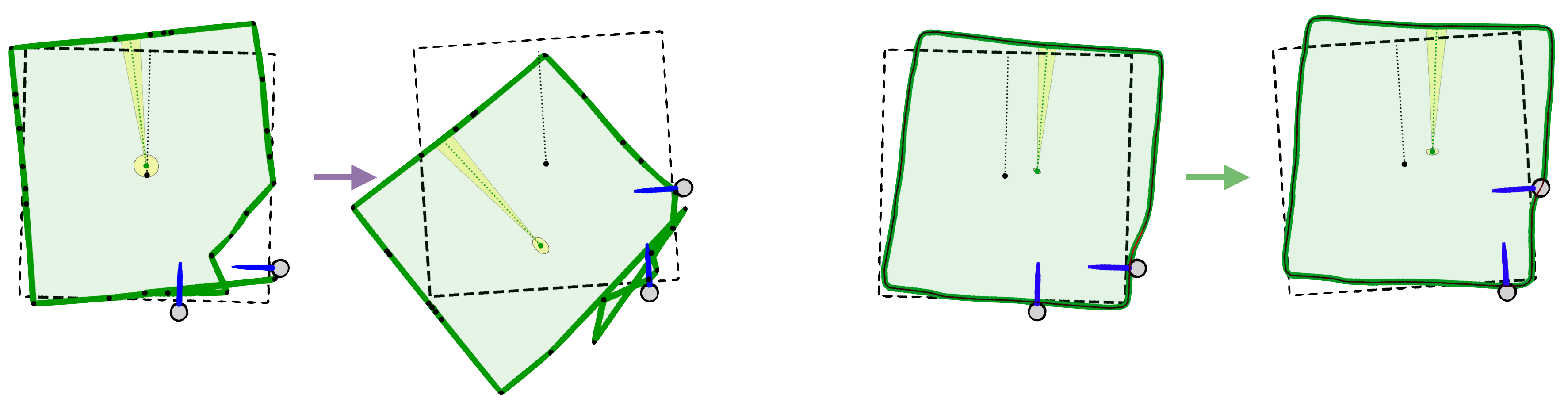}
	\caption{\textbf{[left]} An example of data association failure in the baseline parametric representation~\cite{yu2015shape}. Without discriminating features, committing to vertex associations affects the entire optimization. \textbf{[right]} The GP does not require associations, and the kernel smooths out outlier measurements.}
	\label{fig:yu_da} 
	\GobbleMedium
\end{figure}
\begin{figure}[!b]
	\GobbleLarge
	\centering
	\includegraphics[width=0.9\columnwidth]{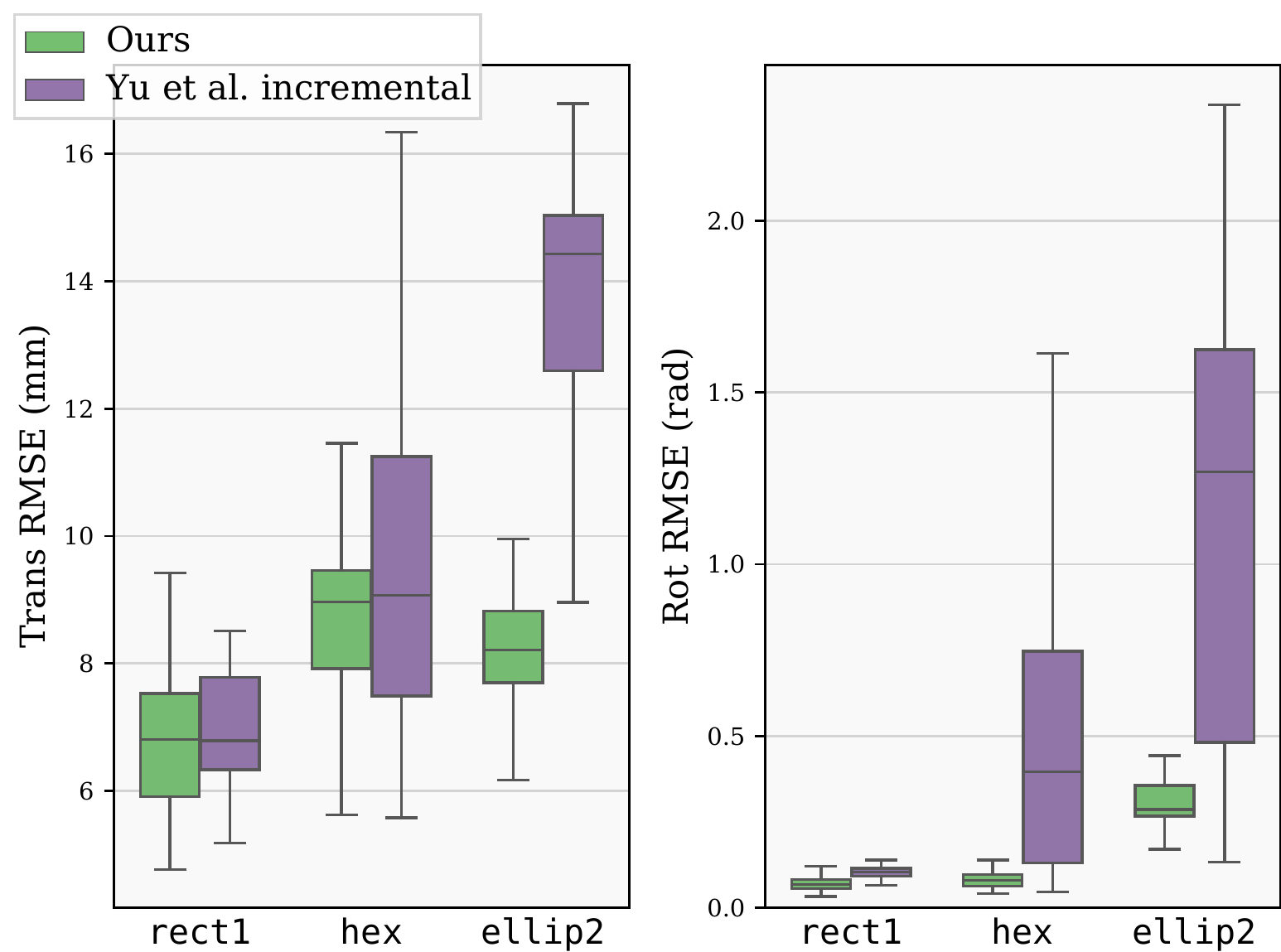}
	\caption{Translation and rotation RMSE box-plots across the $50$ simulation trials, for each of the objects. \textit{Yu \etal~incremental} performs comparably for \texttt{rect1}, but has higher variance and error for the more complex shapes.}
	\label{fig:sim_RMSE} 
	\GobbleMedium
\end{figure}
%

\boldsubheading{Results}
\begin{figure*}
\centering
\begin{minipage}[b]{.71\textwidth}
\includegraphics[width=\textwidth]{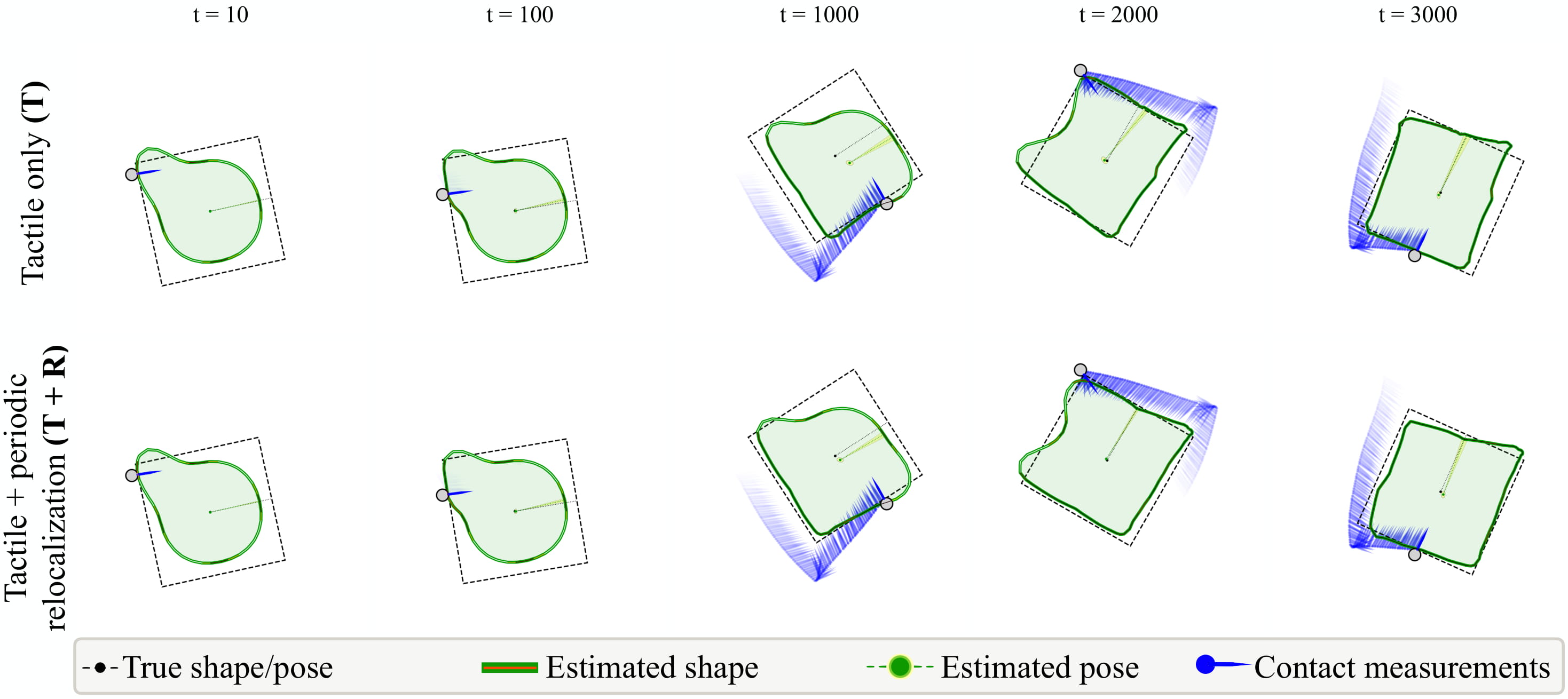}
\caption{Representative results from the real-world estimation task. We compare our tactile only \textbf{(T)} result to one aided by periodic relocalization \textbf{(T + R)}. We add $10$ such events in the trial, and the reduced pose drift improves shape prediction.}\label{fig:real_pose_shape}
\end{minipage}\qquad
\begin{minipage}[b]{.25\textwidth}
\includegraphics[width=\textwidth]{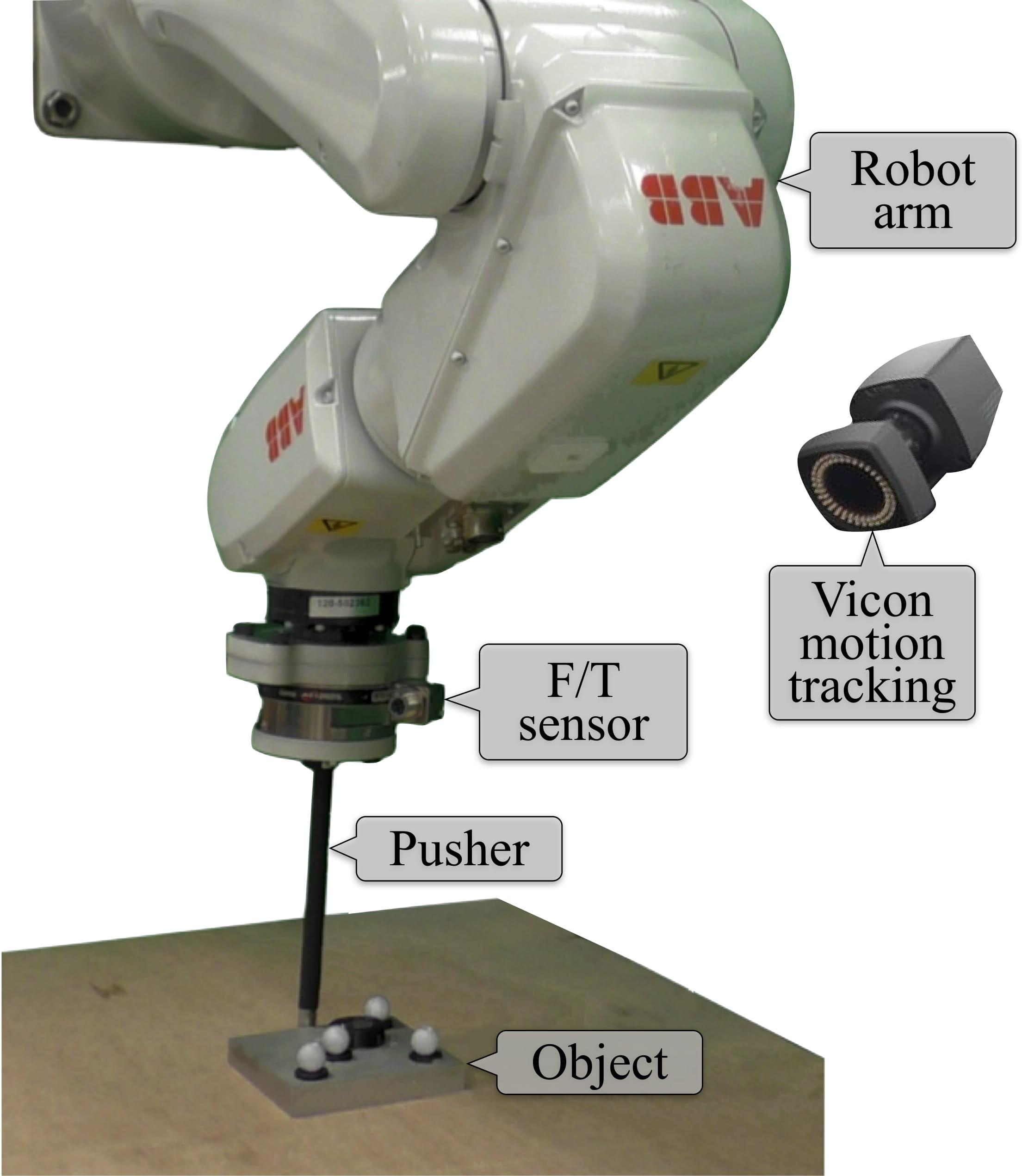}
\caption{Our data collection setup: The ABB IRB 120, with F/T sensing, pushes the block on a plywood surface. The Vicon system tracks the object as ground-truth in our evaluation.}\label{fig:real_data}
\end{minipage}
\GobbleLarge
\end{figure*}
We highlight the qualitative results of a few representative trials in Fig. \ref{fig:sim_pose_shape}. We observe the evolution of object shape from the initial circular prior to the final shape, with pose estimates that match well with ground-truth. 

Fig. \ref{fig:sim_MHD} shows the decreasing MHD shape error over the $50$ trials. The uncertainty threshold of the GPIS $\sigma_{\text{thresh}}^2$ (Section \ref{ssec:gpis_2}) prevents shape updates over repeated exploration, and hence the curve flattens out. This trades-off accuracy for speed, but we find little perceivable difference in the final models. The baseline has larger error for \texttt{ellip2}, a shape which their formulation cannot easily represent. Moreover, the uncertainty of the shape estimates are high, due to data association failures. Our representation has no explicit correspondences, and the kernel smooths out outlier measurements. An example of these effects are seen in Fig.~\ref{fig:yu_da}. A similar trend is seen in pose RMSE across all trials (Fig. \ref{fig:sim_RMSE}). The baseline shows comparable performance only with \texttt{rect1}, as the shape control points can easily represent it.  

Finally, we quantify the computational impact of both the local GPIS regression and incremental fixed-lag optimizer (Fig. \ref{fig:timing_plot}). Fixed-lag smoothing keeps the optimization time bounded, and prevents linear complexity rise. Spatial partitioning keeps online query time low, and $\sigma_{\text{thresh}}^2$ results in less frequent updates over time for both methods. The combination of these two give us an average compute time of about $10~\text{ms}$ or $100~\text{Hz}$. For reference, at $60$~mm/s, that equates to an end-effector travel distance of $0.6~\text{mm}$ per computation. The maximum time taken by a re-linearization step is $55~\text{ms}$  ($3.3~\text{mm}$ travel).
\begin{figure}[b]
	\GobbleMedium
	\centering
	\includegraphics[width=0.9\columnwidth]{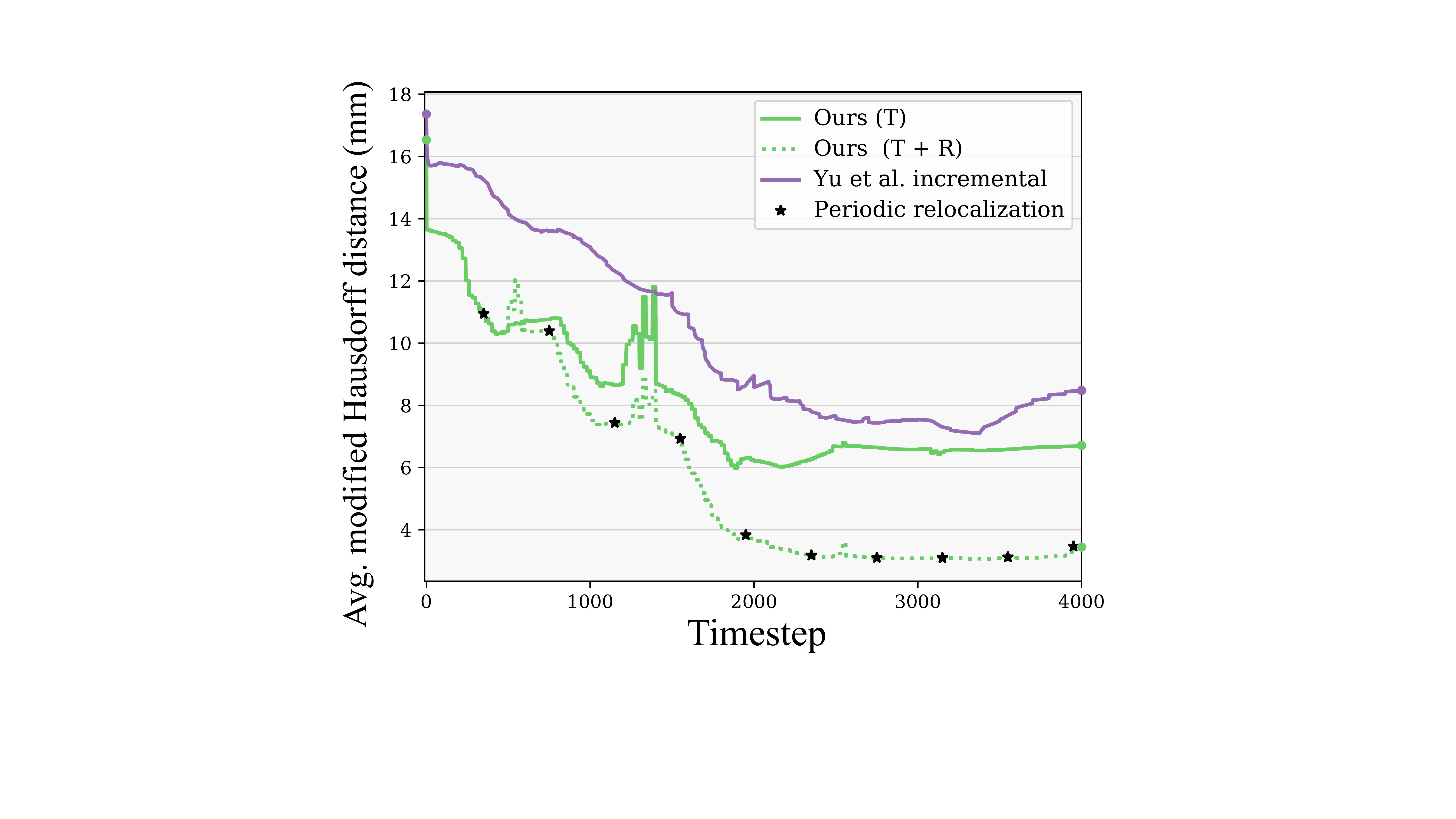}
	\caption{Average MHD w.r.t. the ground-truth  model for the real-world experiments, compared against the baseline. With just a few relocalization events, we can achieve far lower shape error.}
	\label{fig:MHD_plot_real} 
	\GobbleMedium
\end{figure}
%
\subsection{Real-world tactile exploration}
\label{ssec:expts_2}
\boldsubheading{Setup}
We carry out an identical tactile exploration task with the pusher-slider setup in Fig. \ref{fig:real_data}. An ABB IRB 120 industrial robotic arm circumnavigates a square object ($98$~mm side) at the rate of $20$~mm/s. We perform the experiments on a plywood surface, with object-pusher and object-surface coefficient of friction both $\approx 0.25$. We use a single cylindrical rod with a rigidly attached ATI F/T sensor that measures reaction force. Contact is detected with a force threshold, set conservatively to reduce the effect of measurement noise. Ground-truth is collected with Vicon, tracking reflective markers on the object. 

We collect three trials of $4000$ timesteps each, with tactile measurements and ground-truth. In this case, we do not record force magnitude, but only contact normals. We instead map the pusher velocity to forces via the motion cone, and reasoning about sticking and slipping~\cite{mason1986mechanics}. 

\boldsubheading{Results}
The top row \textbf{(T)} of Fig. \ref{fig:real_pose_shape} shows the evolution of shape and pose over the runtime. When compared to the simulation results (Fig. \ref{fig:sim_pose_shape}), we notice aberrations in the global shape. We can attribute these to: \textbf{(i)} lack of a second pusher, which enables better localization and stable pushes, and \textbf{(ii)} motion model uncertainty in real-world pushing. 

The bottom row (T + R) of Fig. \ref{fig:real_pose_shape} describes an additional scenario, where we demonstrate better results when we can periodically relocalize the object. This is a proxy for combining noisy global estimates from vision in a difficult perception task with large occlusions. To illustrate this, we crudely simulate $10$ such events over each trial using global Vicon measurements with Gaussian noise. 

Fig. \ref{fig:MHD_plot_real} plots the evolution of shape error over the three trials. Decrease in shape error is correlated to relocalization events, highlighting the importance of reducing pose drift. Finally, Table \ref{tab:rmse_real} shows the RMSE of our experiments. 
\begin{table}[t]
\GobbleSmall
\centering
\caption{RMSE for real-world tactile exploration. Apart from \textit{Yu \etal~incremental}, we also compare to a method aided by periodic relocalization.}
\label{tab:rmse_real}
\resizebox{0.9\columnwidth}{!}{%
\begin{tabular}{@{}ccc@{}}
\toprule
Method                 & Trans. RMSE (mm) & Rot. RMSE (rad) \\ \midrule
\textbf{Ours (T)} & 10.60 ± 2.74     & 0.09 ± 0.02     \\
Ours (T + R)           & 4.60 ± 1.00      & 0.09 ± 0.01     \\
Yu et al. incremental  & 12.75 ± 4.01    & 0.17 ± 0.03               \\ \bottomrule
\end{tabular}%
}
\GobbleLarge
\end{table}
%
\section{Conclusion}
\label{sec:conc}
We formulate a method for estimating shape and pose of a planar object from a stream of tactile measurements. The GPIS reconstructs the object shape, while geometry and physics-based constraints optimize for pose. By alternating between these steps, we show real-time tactile SLAM in both simulated and real-world settings. This method can potentially accommodate tactile arrays and vision, and be extended beyond planar pushing.


In the future, we wish to build on this framework for online SLAM with dense sensors, like the GelSight~\cite{yuan2017gelsight} or GelSlim~\cite{donlon2018gelslim}, to reconstruct complex 3-D objects. Multi-hypothesis inference methods~\cite{hsiao2019mh} can relax our assumption of known initial pose, and learned shape priors \cite{wang20183d} can better initialize unknown objects. Knowledge about posterior uncertainty can guide active exploration~\cite{dragiev2013uncertainty}, and perform uncertainty-reducing actions~\cite{dogar2012planning}. 

\footnotesize
\bibliographystyle{ieeetr}
\bibliography{references}
\end{document}